\documentclass[runningheads]{llncs}

\usepackage{eccv}

\usepackage{amsmath,amsfonts,amssymb}
\usepackage{algorithmic}
\usepackage{algorithm}
\usepackage{array}
\usepackage{textcomp}
\usepackage{url}
\usepackage{verbatim}
\usepackage{graphicx}
\usepackage{cite}
\usepackage[breaklinks,colorlinks,pagebackref=true,
	breaklinks=true,
	colorlinks=false,
	bookmarks=false,
    linkcolor=red,           
    urlcolor=magenta,        
]{hyperref}

\usepackage{multirow}
\usepackage{xcolor}
\usepackage{colortbl}
\usepackage{threeparttable}
\usepackage{makecell}
\usepackage{adjustbox}
\usepackage{stfloats}
\usepackage{tabularray}

\newcolumntype{C}{>{\centering\arraybackslash}m{1.4cm}}
\newcolumntype{L}{>{\raggedright\arraybackslash}m{2.5cm}}
\newcolumntype{R}{>{\raggedleft\arraybackslash}m{1.5cm}}
\usepackage{siunitx}

\usepackage{microtype}

\usepackage{afterpage}
\usepackage[table]{xcolor}
\usepackage{tabularx}
\usepackage{booktabs}
\usepackage{tabularx}
\usepackage{array}
\newcolumntype{C}{>{\centering\arraybackslash}X}
\newlength{\oldaboverulesep}
\newlength{\oldbelowrulesep}

\usepackage{hhline}
\usepackage{eccvabbrv}

\usepackage{graphicx}
\usepackage{booktabs}

\usepackage[accsupp]{axessibility}  

\usepackage{orcidlink}
\begin{document}

\title{Towards Open-World Referring Expression Comprehension: A Benchmark with Training-free Multi-task Consistency Checker} 

\titlerunning{OpenRef: Open-World REC Benchmark and Multi-task Consistency Checker}

\author{Zongjian Wu\inst{1}\orcidlink{0009-0008-5590-9998} \and
Lei Zhang\inst{1}\thanks{Corresponding author}\orcidlink{0000-0002-5305-8543}}

\authorrunning{Z.~Wu and L.~Zhang}

\institute{School of Microelectronics and Communication Engineering, Chongqing University, Chongqing 400044, China\\
\email{zongjianwu@stu.cqu.edu.cn, leizhang@cqu.edu.cn}}

\maketitle

\begin{abstract}
  Referring expression comprehension (REC) aims to localize a target object within an image based on a given expression. Although recent advances in vision-language models have led to substantial improvements in REC tasks, current REC benchmarks often hold simple scenarios and the assumption that each expression maps to a unique object. These limitations hinder the deployment of REC models in open-world environments. To fill this gap, we introduce OpenRef, a new benchmark for REC in complex visual and linguistic scenarios. OpenRef features three key advancements: 1) Diverse visual scenarios: spanning diverse visual domains, including ground views, drone views, dark scenes and adverse weather conditions; 2) Variable target counts: breaking the single-target limitation with multi-target and none-target samples; 3) Rich vocabulary types: incorporating proper nouns, polysemous words and ordinal terms to fit a wider range of expression needs. Furthermore, as traditional metrics are insufficient for open-world setting, we leverage F1 to measure grounding accuracy and propose N3R (Negative Relative Rejection Reliability) to assess relative rejection reliability against negative expressions. Finally, we introduce Multi-task Consistency Checker (MCC), a training-free but plug-and-play strategy that enhances model performance with one click by enforcing consistency self-verification. 
  Extensive experiments demonstrate that this work significantly advances the performance of existing REC models in complex scenarios, paving the way for open-world REC. Project page: \url{https://zongjianwu.github.io/openref}
\end{abstract}

\begin{figure}[t]
	\centering
	\includegraphics[width=\linewidth]{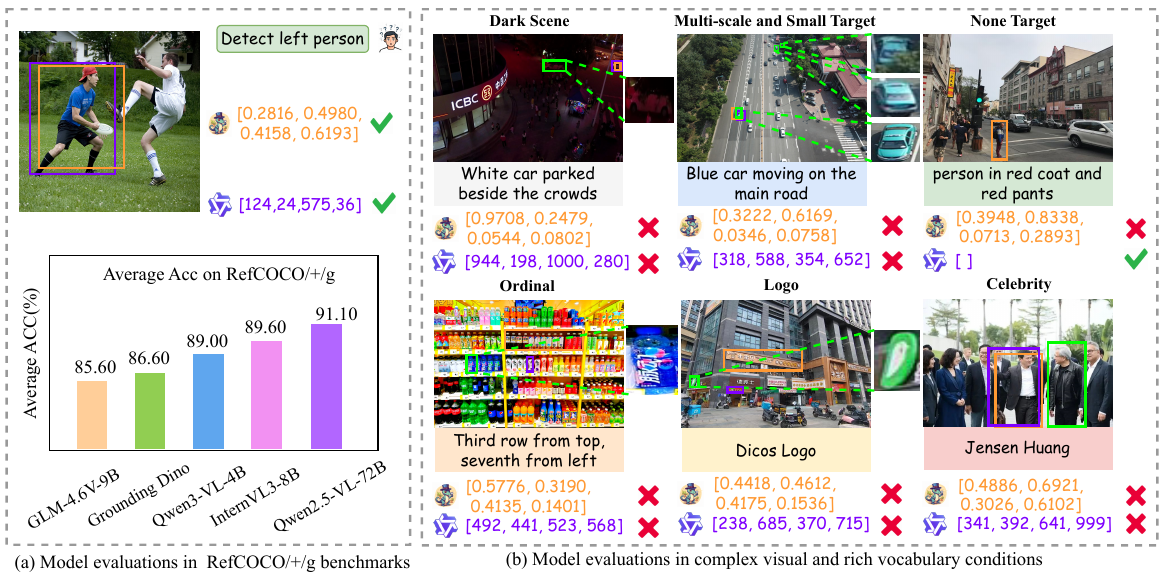}
	\caption{Comparison of RefCOCO and the proposed \textbf{OpenRef} benchmark. (a) Existing REC benchmarks focus on images with simple scenarios, objects occupying entire image and single-target, leading to performance saturation of the REC models. (b) The proposed OpenRef, a comprehensive benchmark across different dimensions, which unveils the evils of current SOTA models in REC tasks and sheds light on new insights for practical deployment of the model in open scenarios.}
	\label{fig: motivation of the benchmark}
\end{figure}

\section{Introduction}
\label{sec:intro}
Referring expression comprehension (REC) aims to locate the visual target in an image guided by the given expression. REC connects image regions with natural language, which contributes to downstream vision-language tasks such as image captioning\cite{rasheed2024glamm,jiang2022visual,peng2025patch,mao2022rethinking}, image editing\cite{liu2024referring,pathiraja2025refedit,xu2025insightedit} and embodied intelligence\cite{fan2023aerial, honerkamp2024language, huang2025roboground, jiang2025robot, liu2023aerialvln}. RefCOCO/+/g\cite{mao2016generation,yu2016modeling} are three most popular benchmarks for REC, providing referring expressions for images in MS-COCO\cite{lin2014microsoft} dataset, many previous works\cite{2stageREC_kazemzadeh2014referitgame,2stageREC_rohrbach2016grounding,2stageREC_wang2019neighbourhood,2stageREC_yang2019dynamic,2stageREC_yu2018mattnet,2stageREC_zhang2018grounding,oneREC_dai2025multi,oneREC_kamath2021mdetr,oneREC_xiao2024hivg,oneREC_xiao2024oneref,oneREC_zheng2025look,oneREC_deng2021transvg,mllm_kosmos2,mllm_internvl2_5} have proposed various methods and achieved impressive results on these benchmarks. 

\begin{figure*}[t]
	\centering
	\includegraphics[width=\textwidth]{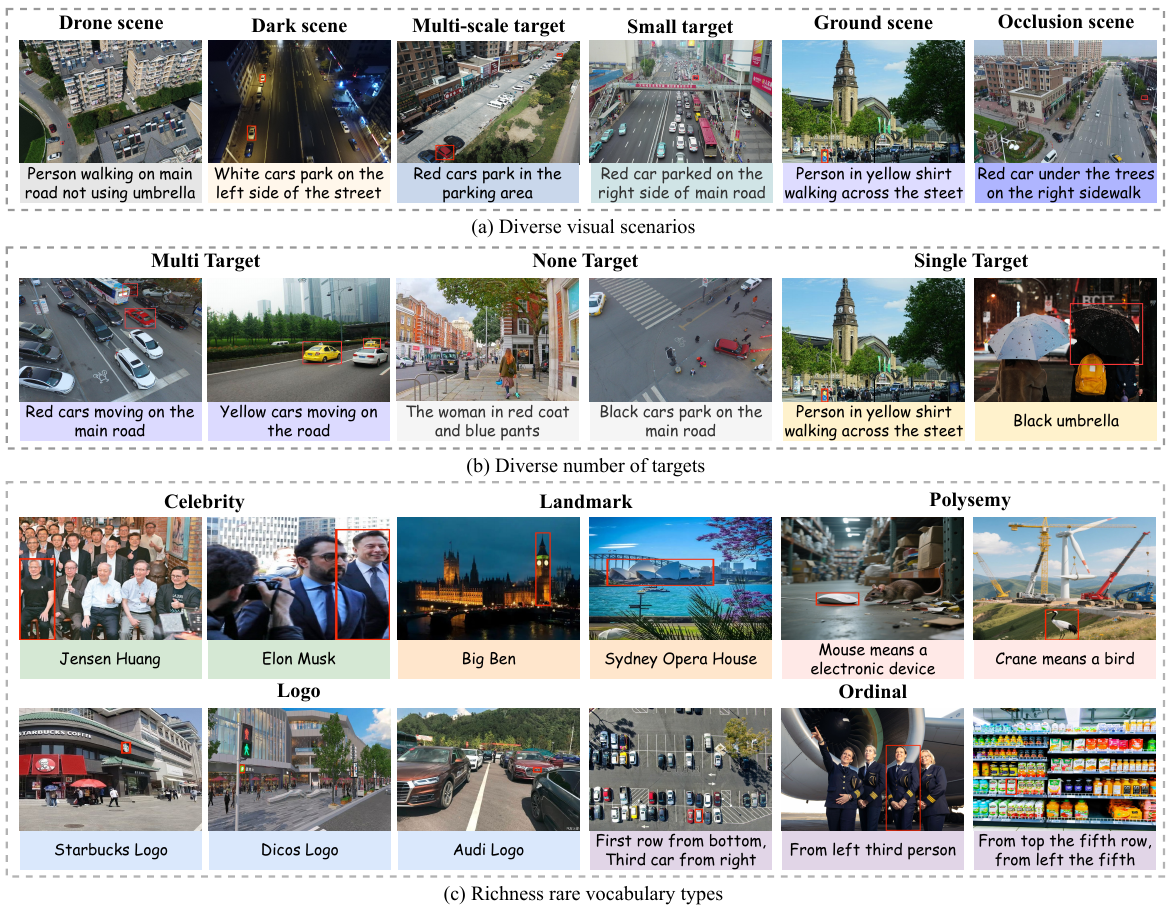}
	\caption{Examples of various challenges. (a) OpenRef includes diverse visual scenarios, providing a real-world setting for REC models. (b) OpenRef includes variable target counts, covering none-target, single-target and multi-target samples. (c) OpenRef includes rich vocabulary types to test cross-modal understanding abilities of REC models.}
	\label{fig:data_vis}
\end{figure*}

However, these benchmarks are typically constructed under controlled, simplistic settings, failing to capture the dynamic and open-world nature of real-world scenarios. As illustrated in \cref{fig: motivation of the benchmark}, REC models that perform well on RefCOCO exhibit a substantial performance drop when evaluated under more complex real-world scenarios. This exposes that current benchmarks have three key limitations: 1) Simplified visual scenarios: lacking various lighting, weather, small targets, occlusion and other complex conditions in real world; 2) Fixed number of referring targets: based on the assumption that one expression corresponds to only one target, neglecting none-target and multi-target scenarios; 3) Restricted vocabulary diversity: insufficient coverage of proper nouns, ordinal terms and polysemous words. These overlooked problems by existing benchmarks have resulted in the closure of the existing referring expression comprehension paradigms, making it difficult to deploy in open-world scenarios. To solve these problems, we propose \textbf{OpenRef}, an open-world referring expression comprehension benchmark, as is shown in \cref{fig:data_vis}, which has three key characteristics. 1) Diverse visual scenarios: spanning diverse visual domains, including ground views, drone views, dark scenes and adverse weather conditions; 2) Variable target counts: breaking the single-target limitation with multi-target and none-target samples; 3) Rich vocabulary types: incorporating proper nouns, polysemous words and ordinal terms to meet a wider range of expressive needs. 

Moreover, traditional Referring Expression Comprehension protocols are largely restricted to single-target scenarios, relying on binary Intersection-over-Union (IoU) thresholds that fail to generalize to complex, multi-target environments. To bridge this gap, we reformulate the evaluation framework by adopting an F1-based metric, which facilitates a balanced assessment of instance retrieval and hallucination suppression. Furthermore, existing absolute metrics (e.g., binary rejection success) in negative expression scenarios often suffer from ranking ambiguity, failing to differentiate between models that vary in error severity, such as those with different box counts or confidence distributions. Such metrics cannot truly reflect model reliability based on relative performance in rejecting negative expressions. To address these limitations, we introduce two synergistic metrics: F1 and N3R (Negative Relative Rejection Reliability). F1 jointly evaluates precision and recall to penalize hallucinations across both single-target and multi-target scenarios. N3R quantifies a model's relative capacity to reject negative expressions, which is still an unresolved issue. These metrics shift REC evaluation from simple geometric overlap to a rigorous assessment of semantic grounding and instance-level reliability.

Last but not least, we propose a crucial Multi-task Consistency Checker (MCC), which is a training-free but plug-and-play inference strategy designed to harmonize multimodal comprehension and substantially improve existing REC models with one-click. Our key insight is that MLLMs frequently exhibit contradictions in complex scenarios: the number of objects identified via referring counting often conflicts with the number of bounding boxes generated via REC. To transform this conflict into a synergistic advantage, without re-training models, MCC pursues inherent multi-task consistency to improve performance. When an inconsistency is detected, MCC triggers a self-verification mechanism, forcing model to re-examine visual context, effectively rectifying false positives and refining grounding precision. By enforcing multi-task consensus at inference for any REC model, a substantial performance gain is achieved for open-world REC.

Our main contributions are summarized as follows:
\begin{itemize}
    \item \textbf{Open-World REC Benchmark}: We propose OpenRef designed for complex visual and linguistic scenarios, shifting REC towards open-world Referring Expression Comprehension.
    \item \textbf{Synergistic Evaluation Metrics:} We introduce holistic evaluation protocols in open-world scenarios, including F1 to measure grounding accuracy and N3R to assess relative rejection reliability.
    \item \textbf{Multi-task Consistency Checker:} We develop a training-free strategy by enforcing multi-task consistency verification to improve any REC model performance under open-world REC scenarios with one-click.
\end{itemize}

\section{Related Works}
\subsection{Conventional REC Benchmarks}
Traditional benchmarks such as RefCOCO/+/g\cite{mao2016generation,yu2016modeling} operate under a closed-world assumption that each expression maps to a unique target. Recent efforts have relaxed this assumption by introducing multi-target and none-target scenarios. G-RES\cite{opendata_liu2023gres} first broadened task formulation to open-world settings. Two complementary lines of research have emerged to improve real-world applicability in REC. One line focuses on increasing linguistic and relational reasoning demands. ReMeREC\cite{opendata_hu2025remerec} emphasized complex inter-entity relations, forcing models to perform deeper multimodal parsing rather than rely on shallow cues. Ref-Adv\cite{opendata_akula2020words}, Cops-Ref\cite{openda_chen2020cops}, Ref-Reasoning\cite{openda_yang2020graph} and FineCops-Ref~\cite{opendata_liu2024finecops} established a benchmark with meticulously edited negative expressions or images to expose superficial keyword matching.  OVDEval\cite{opendata_yao2024evaluate} demonstrated importance of negative nouns for evaluating open-vocabulary detection. Despite these advances, most reasoning oriented benchmarks are built upon previous benchmarks \cite{lin2014microsoft,plummer2015flickr30k,2stageREC_kazemzadeh2014referitgame,trad_wu2020phrasecut}. As a result, they inherit object-centric visual biases and remain limited to simple scenarios. Another line mainly considers visual scene changes. RSVG\cite{opendata_zhan2023rsvg}, Ref-Drone\cite{opendata_sun2025refdrone} and SOREC\cite{opendata_goto2025referring} partially addressed this gap by introducing diverse visual scenarios, but lack carefully edited negative expressions. Moreover, existing works are constrained by limited lexical types and unreasonable evaluation protocols, hindering a comprehensive and rigorous assessment of REC models. To address these problems, we introduce \textbf{OpenRef} towards open-world REC.

\subsection{Specialist-based REC Methods}
Early specialist models~\cite{2stageREC_kazemzadeh2014referitgame, 2stageREC_rohrbach2016grounding, 2stageREC_zhang2018grounding, 2stageREC_yu2018mattnet, 2stageREC_yang2019dynamic, 2stageREC_wang2019neighbourhood} typically employed pre-trained detectors to generate region proposals and localized targets through text-to-region matching, they are often limited by the performance of pre-trained detectors. Subsequent methods transitioned to one-stage design,~\cite{oneREC_yang2019fast} directly regressed bounding boxes from multimodal feature maps to achieve better efficiency and accuracy. Recent methods have reformulated REC in an end-to-end manner by integrating vision and language branch within a unified Transformer framework. MDETR~\cite{oneREC_kamath2021mdetr} pioneered the integration of free-form expression with DETR through an end-to-end framework. TransVG~\cite{oneREC_deng2021transvg} further simplified REC by directly regressing bounding box coordinates within Transformer-based architectures. Grounding DINO~\cite{oneREC_liu2024grounding} extended Transformer-based grounding to open-vocabulary settings. To mitigate parameter redundancy, OneRef~\cite{oneREC_xiao2024oneref} and HiVG~\cite{oneREC_xiao2024hivg} leveraged parameter-efficient fine-tuning strategies to reduce model parameters. C3VG~\cite{oneREC_dai2025multi} proposed a multi-stage framework, which first performs preliminary grounding and further refines coarse coordinates. SSRVG~\cite{oneREC_zheng2025look} focused on structured language understanding by decomposing referring expressions around a center noun and progressively incorporating contextual cues to guide localization. Despite superior performance on standard benchmarks, these REC models remain limited in handling multi-target and none-target cases that frequently arise in  real-world scenarios. This limitation is to be resolved in our OpenRef.

\subsection{MLLM-based REC Methods}
Recent years have witnessed the remarkable success of Multimodal Large Language Models (MLLMs) in REC tasks. Early attempts like Kosmos-2~\cite{mllm_kosmos2} and MiniGPT-V2~\cite{mllm_minigptv2} integrated visual grounding capabilities by treating bounding box coordinates as special textual tokens. Recent architectures GLM-4.6V\cite{mllm_glm4.5/4.6v_2025} enhanced cross-modal alignment through a high-density visual perception mechanism. Keye-VL-1.5\cite{mllm_keye} introduced a multi-scale feature fusion strategy, enabling the model to capture fine-grained objects. MiniCPM-V-4.5\cite{mllm_minicpmv45} leveraged an adaptive high-definition tiling strategy, which allows the model to process ultra-high-resolution images with remarkable efficiency. CogVLM~\cite{mllm_cogvlm} utilized a dedicated visual expert to decouple visual and linguistic features.  InternVL-2.5~\cite{mllm_internvl2_5} and Qwen2-VL~\cite{mllm_qwen2vl}  utilized massive-scale pre-training and dynamic resolution strategies to improve performance. Moreover, Set-of-Mark (SoM) prompting~\cite{mllm_som} has shown that closed-source models like GPT-4V\cite{mllm_gpt4v} can also achieve extraordinary visual grounding performance. Despite their impressive performance, MLLMs often struggle in none-target and multi-target scenarios. In this work, we introduce OpenRef along with refined evaluation metrics to comprehensively assess reliability and mitigate hallucination-induced overestimation.

\section{Proposed Benchmark: OpenRef}

\subsection{Benchmark Construction}
\begin{figure*}[t]
	\centering
	\includegraphics[width=\textwidth]{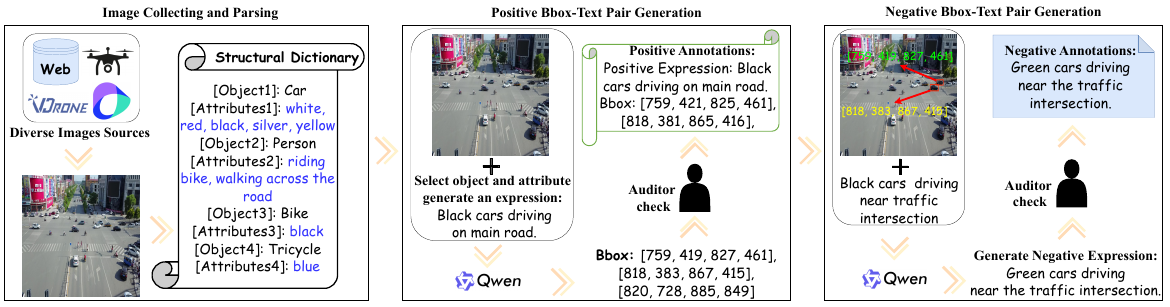}
	\caption{Annotation pipeline to construct OpenRef. We first collect image-text pairs and utilize Qwen3-VL to generate bounding boxes, then generate negative expressions by carefully perturbing positive expressions. This process yields final annotations comprising images, bounding boxes, positive expressions and negative expressions.}
	\label{fig:data_pipeline}
\end{figure*}

To generate high-quality query–bounding box pairs, we propose a semi-automated annotation pipeline as depicted in \cref{fig:data_pipeline}, which comprises three steps.

\textbf{Image Collecting and Parsing.} We have curated a high-quality and diverse visual corpus by combining large-scale web-crawled imagery with specialized drone-view scenes from VisDrone \cite{du2019visdrone}. To address the data scarcity of special linguistic concepts (e.g., polysemous words and ordinal terms), we leverage MLLM to synthesize images. Synergy between realistic in-the-wild captures and targeted synthetic generation ensures both visual complexity and semantic granularity. For each image, we use Qwen3-VL\cite{mllm_qwen3-vl} to generate a structured property dictionary encoding the visual elements, including the object category, inherent attributes and interactive relationships with other objects. Then we use the above information to randomly generate a referring expression.

\textbf{Positive Bbox-Text Pair Generation}: We use Qwen3-VL\cite{mllm_qwen3-vl} to generate corresponding coordinates by leveraging referring expression as a text prompt. Due to the diversity and complexity of visual scenes in the dataset, even the most advanced Qwen3-VL\cite{mllm_qwen3-vl} also has shortcomings, so we manually review and correct the generated coordinates. By correcting inaccurate coordinates and referring expressions in RefDrone~\cite{opendata_sun2025refdrone}, we obtain a small set of additional annotations. All the above parts collectively obtain the positive bbox-text annotations.

\textbf{Negative Bbox-Text Pair Generation}: To generate high-quality negative samples, we employ MLLM to perform fine-grained semantic perturbations on positive referring expressions. Specifically, we systematically manipulate key attributes—including colors, spatial orientations, category nouns and proper nouns to generate challenging descriptions that no longer correspond to any target object within the image. Following this automated generation, a rigorous manual inspection is conducted to filter out ambiguities, ensuring each image is annotated with a precise triplet: a positive expression, a hard negative expression and ground-truth coordinates.

\begin{figure*}[t]
	\centering
	\begin{subfigure}[b]{0.24\textwidth}
		\centering
		\includegraphics[width=\linewidth]{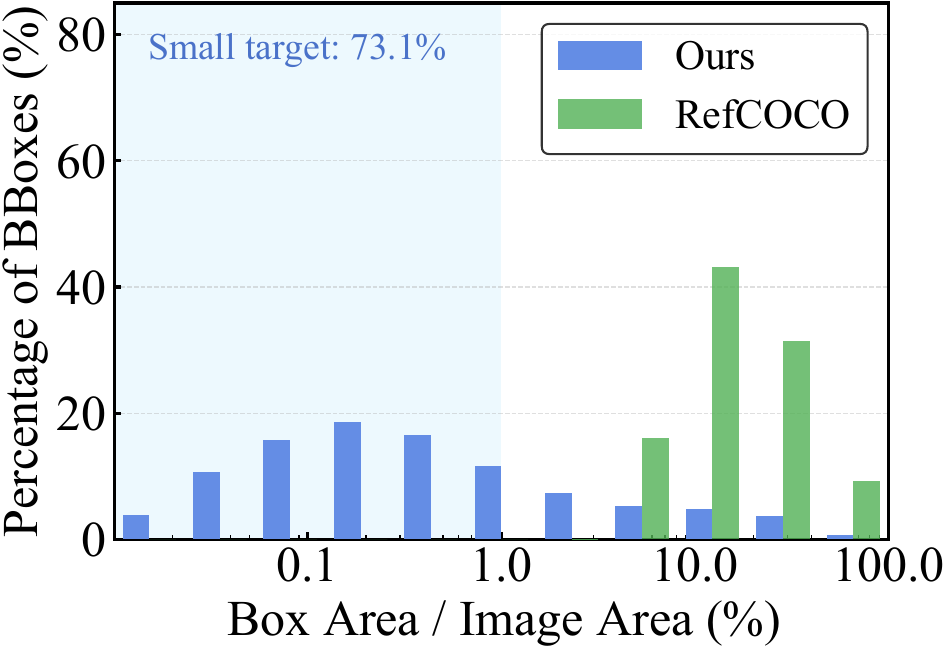}
		\caption{Referring target size distribution}
		\label{fig:sub1}
	\end{subfigure}\hfill 
	\begin{subfigure}[b]{0.24\textwidth}
		\centering
		\includegraphics[width=\linewidth]{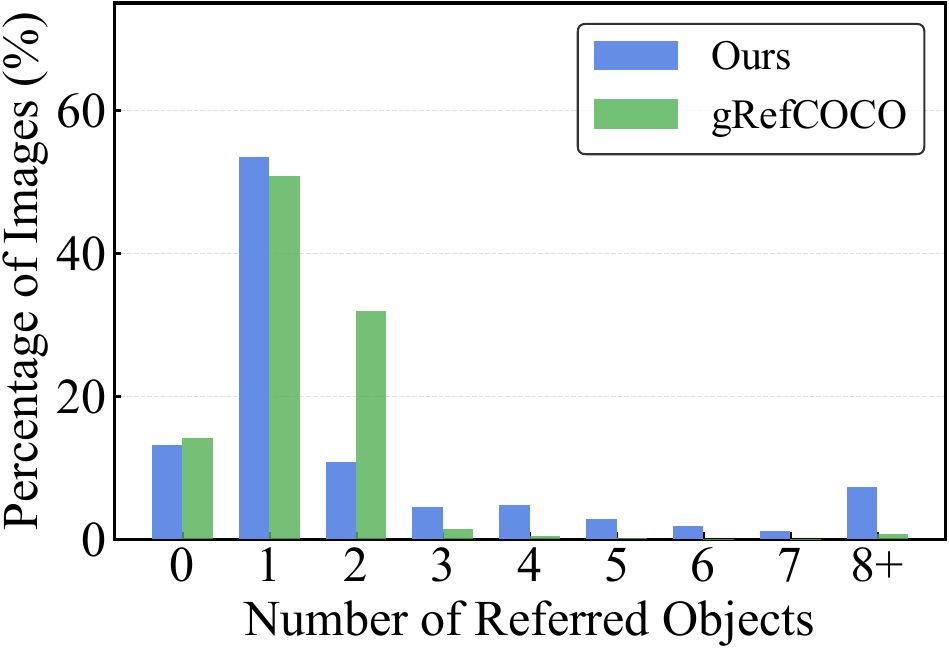}
		\caption{Number of targets per expression}
		\label{fig:sub3}
	\end{subfigure}\hfill
	\begin{subfigure}[b]{0.24\textwidth}
		\centering
		\includegraphics[width=\linewidth]{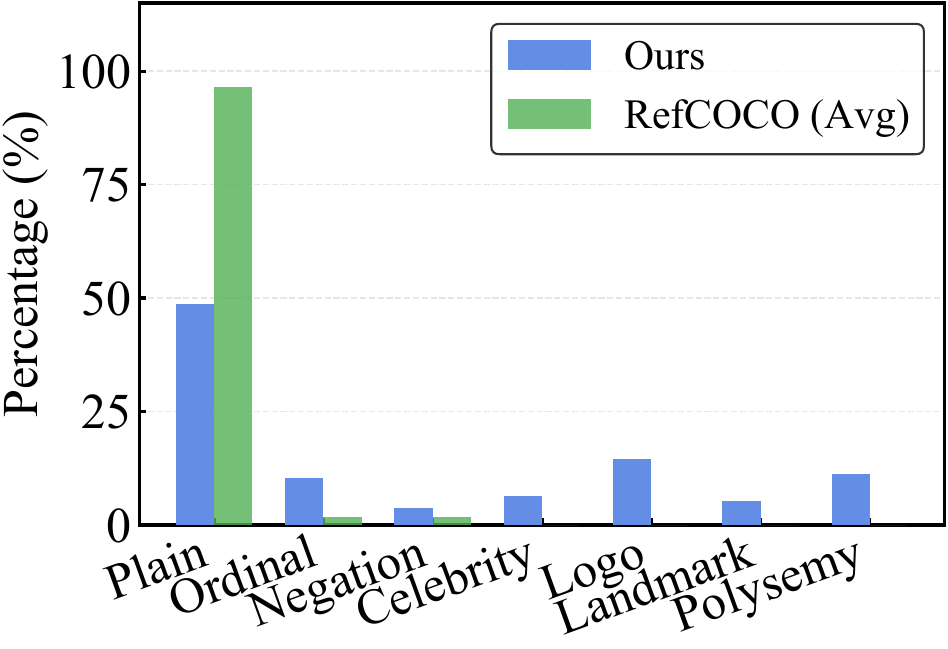}
		\caption{Rare Vocabulary distribution}
		\label{fig:sub2}
	\end{subfigure}\hfill
	\begin{subfigure}[b]{0.232\textwidth}
		\centering
		\includegraphics[width=\linewidth]{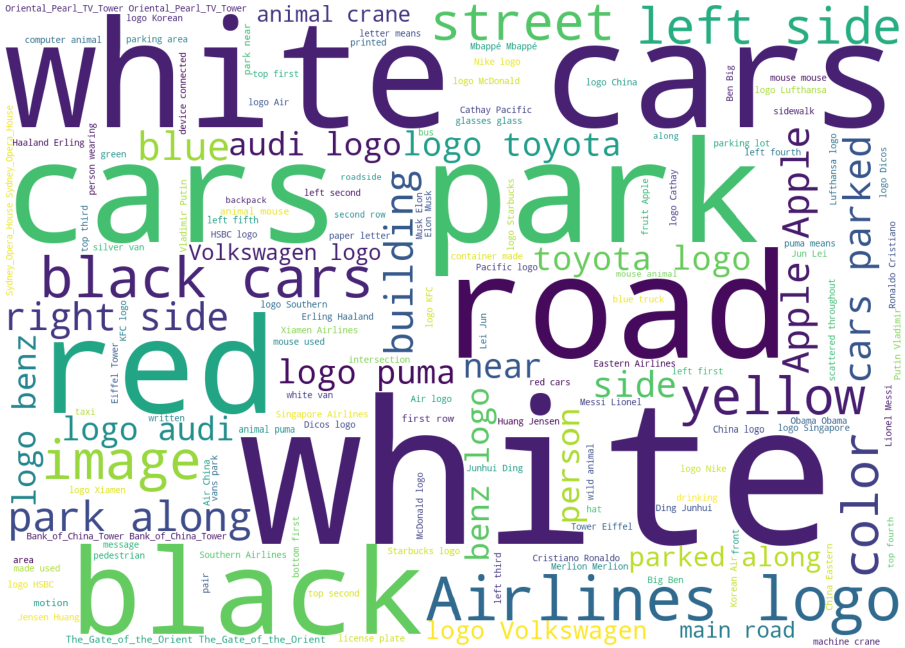}
		\caption{Word cloud of positive expressions}
		\label{fig:sub4}
	\end{subfigure}
	
	\caption{Visual and linguistic statistics of OpenRef. Compared to traditional benchmarks, OpenRef demonstrates remarkable advantages in handling small-target and multi-target scenarios, as well as in handling rare vocabulary distributions.}
	\label{fig:four_images}
\end{figure*}

\subsection{Benchmark Analysis}
OpenRef contains 32,735 referring expressions across 17,586 images, comprising 37698 object instances. \cref{fig:four_images} highlights the three key challenges in OpenRef.

\textbf{1) Small-scale and Multi-scale Challenges.} \cref{fig:four_images} (a) illustrates scale distribution in OpenRef, characterized by a high density of small-scale instances and the coexistence of multiple scales within a single scene. We categorize target scales into four tiers according to the relative area occupancy of their bounding boxes within the image, including \textit{tiny} ($<$ 1\%), \textit{small} (1\%--10\%), \textit{medium} (10\%--50\%) and \textit{large} ($>$ 50\%). Statistical analysis reveals a significant prevalence of small-scale targets, with \textit{tiny} ($<$ 1\%) alone accounting for 73.1\% of all targets, posing a substantial challenge for REC in open-world environment.

\textbf{2) Multi-target and None-target Samples.} Unlike traditional benchmarks that primarily focus on single-target scenarios, OpenRef includes more multi-target and none-target scenarios.  \Cref{fig:four_images} (b) illustrates the target counts distribution, revealing higher complexity in multi-target scenarios compared to gRefCOCO\cite{opendata_liu2023gres}, where a majority of expressions refer to only one or two targets.

\textbf{3) Rich Vocabulary Types.} Traditional benchmarks RefCOCO/+/g \cite{mao2016generation,yu2016modeling} mainly rely on simplified referring expressions, lacking diverse vocabulary types. In contrast, OpenRef introduces a significantly greater lexical diversity. As illustrated in \cref{fig:four_images} (c), OpenRef provides a broader distribution of rare vocabulary types compared to existing benchmarks, demonstrating remarkable advantage in linguistic complexity. Furthermore, word cloud in \cref{fig:four_images} (d) exemplifies a wide spectrum of linguistic constructs incorporated into expressions, encompassing proper nouns, ordinals and polysemous terms. This diversity compels REC models to progress from basic object recognition to sophisticated semantic comprehension.

\begin{table*}[t]
	\centering
	\small
	\fontsize{7pt}{7.5pt}\selectfont
	\setlength{\tabcolsep}{1.5pt} 
	\caption{Comparison of REC benchmarks. 
		\label{tab:benchmark_comparison}
		\textbf{Tar. Mul.}: target multiplicity, refers to whether include single-target and multi-target scenarios.; \textbf{Neg. Exp.}: negative expression refers to non-existent target; \textbf{Pers. Div.}: perspective diversity, covers multi-view
		scenarios such as ground views and drone views; \textbf{Dom. Rob.}: domain robustness, accounts for challenging environments including low light, small target and adverse weather conditions; \textbf{Lex. Rich.}: lexical richness, such as proper nouns, ordinal numbers; \textbf{Hallu. Ass.}: hallucination assessment, evaluates model's ability on penalizing false positive predictions; \textbf{Rel. Rank.}: negative relative rejection ranking.}
	
	\begin{tabularx}{\textwidth}{l l *{7}{>{\centering\arraybackslash}X}}
		\toprule
		\multirow{3}{*}{\textbf{Benchmark}} & \multirow{3}{*}{\textbf{Venue}} & \multicolumn{5}{c}{\textbf{\makecell[c]{Linguistic \&  Visual Characteristics}}} & \multicolumn{2}{c}{\textbf{\makecell[c]{Evaluation  Protocol}}} \\
		\cmidrule(lr){3-7} \cmidrule(lr){8-9}
		& & \makecell{Tar.\\Mul.} & \makecell{Neg.\\Exp.} & \makecell{Pers.\\Div.} & \makecell{Dom.\\Rob.} & \makecell{Lex.\\Rich.} & \makecell{Hallu.\\Ass.} & \makecell{Rel.\\Rank.} \\
		\midrule
		RefCOCO~\cite{yu2016modeling}    & ECCV'16 &  &  &  &  &  &  &  \\
		RefCOCO+~\cite{yu2016modeling}   & ECCV'16 &  &  &  &  &  &  &  \\
		RefCOCOg~\cite{mao2016generation} & CVPR'16 &  &  &  &  &  &  &  \\
		Cops-Ref~\cite{openda_chen2020cops} & CVPR'20 &  & \checkmark &  &  &  &  &  \\
		Ref-Reasoning~\cite{openda_yang2020graph} & CVPR'20 &  & \checkmark &  &  &  &  &  \\
		Ref-Adv~\cite{opendata_akula2020words} & ACL'21 &  & \checkmark &  &  &  &  &  \\
		gRefCOCO~\cite{opendata_liu2023gres} & CVPR'23 & \checkmark &  &  &  &  &  &  \\
		RSVG~\cite{opendata_zhan2023rsvg} & TGRS'23 &  &  &  &  &  &  &  \\
		Ref-L4~\cite{rasheed2024glamm} & CVPR'24 &  &  &  &  &  &  &  \\
		OVDEval~\cite{opendata_yao2024evaluate} & AAAI'24 &  & \checkmark &  &  &  &  &  \\
		FineCops-Ref~\cite{opendata_yang2025new} & TPAMI'25 &  & \checkmark &  &  &  &  &  \\
		ReMeREC~\cite{opendata_hu2025remerec} & MM'25 &  &  &  &  &  &  &  \\
		SOREC~\cite{opendata_goto2025referring} & ICCV'25 &  &  &  & \checkmark &  &  &  \\
		\midrule
		\textbf{OpenRef (Ours)} &  & \textbf{\checkmark} & \textbf{\checkmark} & \textbf{\checkmark} & \textbf{\checkmark} & \textbf{\checkmark} & \textbf{\checkmark} & \textbf{\checkmark} \\
		\bottomrule
	\end{tabularx}
\end{table*}

\subsection{Benchmark Comparison}
\Cref{tab:benchmark_comparison} compares OpenRef with existing benchmarks. Compared with gRefCOCO\cite{opendata_liu2023gres}, OpenRef presents a broader and more balanced distribution over none-target and multi-target scenarios, better reflecting real-world object multiplicity. In contrast to SOREC\cite{opendata_goto2025referring}, which primarily employs declarative sentences, OpenRef incorporates more diverse vocabulary types, enhancing linguistic richness. Furthermore, OpenRef introduces a more comprehensive evaluation protocol that explicitly accounts for multi-target grounding and none-target rejection, while featuring richer and more challenging visual scenes with diverse lighting conditions, occlusions and small objects. These advantages establish OpenRef as a more realistic and challenging benchmark for open-world REC.

\subsection{Proposed Evaluation Metrics for OpenRef}
To rigorously evaluate REC models across none-target, single-target and multi-target scenarios, we propose two synergistic evaluation metrics, including F1 to measure grounding accuracy and N3R (Negative Relative Rejection Reliability) to assess relative rejection reliability against negative expressions.

\begin{figure}[t]
	\includegraphics[width=\linewidth]{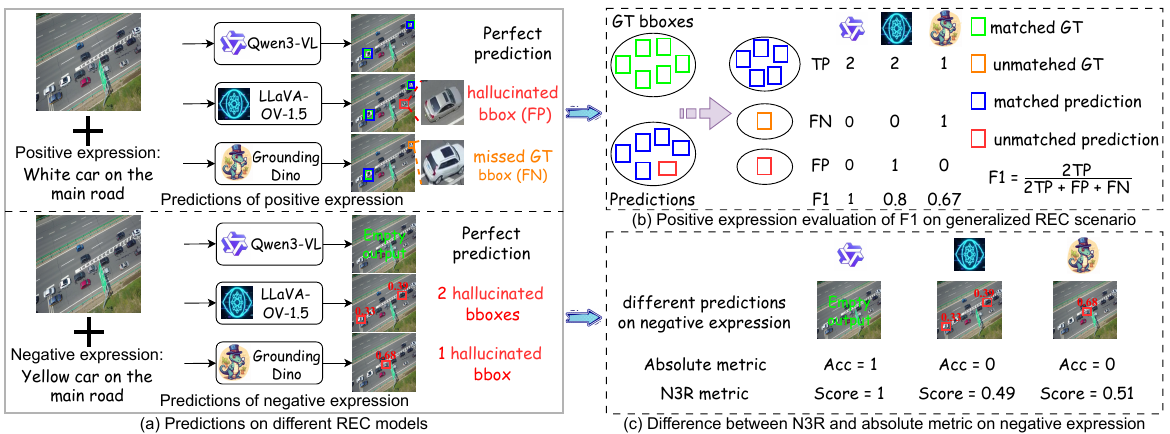}
	\caption{Illustration of the evaluation protocol. (a) Predictions of different models on positive and negative expressions, respectively. Existing models often struggle with hallucinated boxes (false positives, FP) or missed targets (false negatives, FN) in open-world scenarios. (b) F1-score provides a balanced assessment for instance retrieval in open-world environments, and therefore effectively evaluate different models on multi-targets w.r.t. positive expressions. (c) Comparison between N3R and absolute metric. N3R better quantifies relative reliability in rejecting negative expressions, and therefore effectively evaluate the performance of different models w.r.t. negative expressions.}
	\label{fig: metrics}   
\end{figure}

\textbf{Evaluation on Grounding Accuracy.} Unlike traditional REC evaluation protocols that simplify REC task into a single-box retrieval problem (e.g., Top-K Acc), we reformulate generalized REC task in open-world setting through standard F1 score to evaluate grounding accuracy across single-target and multi-target scenarios. F1 naturally penalizes hallucinated bounding boxes and rewards precise multi-instance localization. \Cref{fig: metrics} (b) illustrates the 
calculation process of F1. We define $\mathcal{D} = \{ (I_i, E_i) \}_{i=1}^N$ be a dataset of $N$ image-text pairs. For the $i$-th sample, let $\mathcal{G}_i = \{g_j\}_{j=1}^{|G_i|}$ denote the GT set and $\mathcal{P}_i = \{p_k\}_{k=1}^{|P_i|}$ denote the prediction set. Cardinality $|\mathcal{G}_i|$ can be one (single-target scenario) or greater than one (multi-target scenario). We define the sample-level True Positives ($\text{TP}_i$), False Positives ($\text{FP}_i$) and False Negatives ($\text{FN}_i$), where $\text{TP}_i$ denotes the number of predictions in $\mathcal{P}_i$ that successfully match a GT box in $\mathcal{G}_i$, $\text{FP}_i$ denotes the number of predictions in $\mathcal{P}_i$ that do not have a corresponding GT box (i.e., IoU < 0.5 with all $g_j \in \mathcal{G}_i$) and $\text{FN}_i$ denotes the number of GT boxes in $\mathcal{G}_i$ that are not covered by any prediction (i.e., missed targets). For positive samples (i.e., $|\mathcal{G}_i| > 0$), there is:
$\text{TP}_i = |\mathcal{M}_i|, \quad \text{FP}_i = |\mathcal{P}_i| - |\mathcal{M}_i|, \quad \text{FN}_i = |\mathcal{G}_i| - |\mathcal{M}_i|.$
 We compute Precision and Recall across entire dataset:
\begin{equation}
	Precision = \frac{\sum \text{TP}_i}{\sum (\text{TP}_i + \text{FP}_i)}, \quad Recall = \frac{\sum \text{TP}_i}{\sum (\text{TP}_i + \text{FN}_i)}.
\end{equation}
Then F1 score is computed as:
\begin{equation}
\text{F1} = \frac{1}{N} \sum_{i=1}^N \frac{2 Precision_i\cdot Recall_i}{Precision_i + Recall_i}
= \frac{1}{N} \sum_{i=1}^N \frac{2 \cdot \text{TP}_i}{2 \cdot \text{TP}_i + \text{FP}_i + \text{FN}_i},
\end{equation}

\textbf{Negative Relative Rejection Reliability (N3R).} We introduce N3R to assess the relative rejection reliability of models against negative expressions. (i.e., targets absent from the image). Unlike traditional binary metrics, N3R provides a relative rejection score. \Cref{fig: metrics} (c) illustrates the difference of two metrics: when presented with a negative expression, Grounding DINO and LLaVA-OneVision-1.5 produce different sets of hallucinated boxes, but absolute metric assigns both models the same binary outcome zero, failing to assess the relative rejection reliability of different models. To provide a fine-grained assessment, we implement a confidence-aware scoring mechanism. For MLLMs that do not explicitly output box-level scores, we derive the confidence $c_k$ of the $k$-th box by calculating the mean probability of its constituent coordinate tokens' logits. We calculate reliability score of a negative expression by taking the product of confidence scores associated with each hallucinated box:
\begin{equation}
	\text{N3R} = \frac{1}{N} \sum_{i=1}^{N} \left( \prod_{k=1}^{|\mathcal{P}_i|} (1 - c_k) \right)
\end{equation}
where $c_k \in [0, 1]$ represents the average confidence of the $k$-th spurious box. This multiplicative formulation ensures that high-confidence hallucinations or a large quantity of false positives exponentially diminish the final score, reflecting the intuition that such errors represent a more severe lack of robustness.

\begin{figure}[t]
	\includegraphics[width=\linewidth]{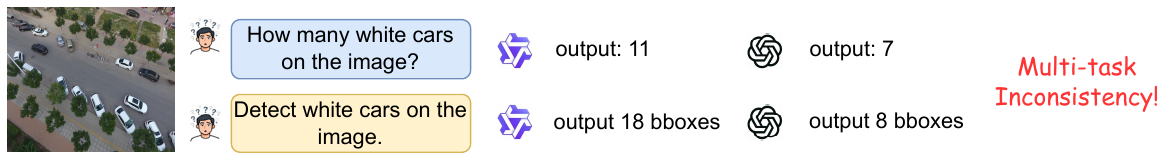}
	\caption{Motivation of MCC. Multi-task inconsistency between REC and Referring counting tasks exists for Qwen3-VL\cite{mllm_qwen3-vl} and GPT\cite{mllm_gpt4v}, respectively. Ensuring the consistency largely helps improve REC, which is the intuition of MCC.}
	\label{fig: MCC motivation}   
\end{figure}
\section{Proposed Multi-task Consistency Checker}
\label{sec:blind}
\begin{figure}[t]
	\includegraphics[width=\linewidth]{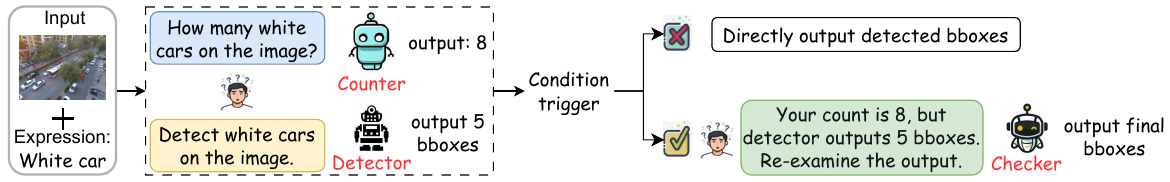}
	\caption{The proposed MCC pipeline with two functions (i.e., counter and detector) and a condition trigger. When there is a task conflict between the counter and detector outputs, the checker re-examine the image and then final results are reported. }
	\label{fig:MCC}   
\end{figure}

We introduce a Multi-task Consistency Checker (MCC) to address multi-target and none-target challenges. Our key insight is that MLLMs frequently exhibit contradictions in different tasks. This training-free but plug-and-play inference strategy serves as a checker that harmonizes latent contradictions between different functional heads of the MLLM. As illustrated in \cref{fig: MCC motivation}, when presented with a text query, the model's numerical prediction $N_{cnt}$ from the counter (i.e., counting task) frequently deviates from the number $N_{rec}$ of instances localized by the detector (i.e., REC task). This logical discrepancy suggests an internal uncertainty in the model’s grounding process. \Cref{fig:MCC} shows the pipeline of MCC. For a given image $I$ and a referring expression $T$, we concurrently execute two distinct tasks: 1) Referring counter predicts total number of target objects, denoted as $N_{cnt} = \mathcal{F}_{RC}(I, T)$ and 2) REC detector predicts a set of bounding boxes, denoted as $\mathcal{B}_{rec} = \{b_1, b_2, \dots, b_k\}$, where $N_{rec} = |\mathcal{B}_{rec}|$. A consistency conflict is defined by the following indicator function $\mathbb{1}_{conflict}$:
\begin{equation}
	\mathbb{1}_{conflict} = \begin{cases} 1, & \text{if } N_{cnt} \neq N_{rec} \\ 0, & \text{if } N_{cnt} = N_{rec} \end{cases}
	\label{eq:conflict_indicator}
\end{equation}
Note that when $\mathbb{1}_{conflict} = 1$, the the trigger is activated, and then MCC initiates a Self-Verification Loop. Rather than accepting the initial disparate outputs, we feed the detected contradiction back into the model as a strategic prompt to derive the final grounded result, formulated as:
\begin{equation}
	\mathcal{B}_{final} = \text{Checker}(I, T, N_{cnt}, \mathcal{B}_{rec})
	\label{eq:arbiter_process}
\end{equation}

The checker is prompted to re-examine the visual context under the explicit premise that a conflict exists, forcing a logical alignment between numerical perception and spatial localization. This "second-pass" reasoning forces the model to re-allocate its attention, either pruning false positive boxes if $N_{rec} > N_{cnt}$ or discovering omitted instances if $N_{cnt} > N_{rec}$. To mitigate the risk of sycophancy hallucination where the model might yield to the arbiter's doubt and erroneously reject all targets, we implement a heuristic fallback. If the checker returns an empty set $\emptyset$ despite both $N_{cnt} > 0$ and $N_{rec} > 0$, the system defaults to the initial $\mathcal{B}_{rec}$ to preserve the \textit{Recall}. This ensures that MCC positively enhances precision without sacrificing the fundamental grounding capability.

\begin{table*}
    \centering
    \fontsize{7}{7.5pt}\selectfont
    \setlength{\tabcolsep}{1pt} 
    \renewcommand{\arraystretch}{1.2}
    \caption{Grounding accuracy of REC models on OpenRef compared with RefCOCO. Our benchmark presents significantly higher difficulty. Best results are in \textbf{bold}.}
    \label{tab:main_results_comparison}
    
    \begin{tabularx}{\textwidth}{@{} l c *{7}{C} c | c @{}}
        \toprule
        \noalign{%
            \global\oldaboverulesep=\aboverulesep
            \global\oldbelowrulesep=\belowrulesep
            \global\aboverulesep=0pt
            \global\belowrulesep=0pt
        }
        
        \multirow{2}{*}{\textbf{Model}} & \multirow{2}{*}{\textbf{Size}} 
        & \multicolumn{5}{c}{\cellcolor{blue!8}\textbf{Lexical Richness}} 
        & \multicolumn{2}{c}{\cellcolor{blue!8}\textbf{Targ. Mul.}} 
        & \cellcolor{blue!8}\textbf{OpenRef}   
        & \textbf{RefCOCO} \\
        
        \cmidrule[0.4pt](lr){3-7} 
        \cmidrule[0.4pt](lr){8-9}
        
        & & 
        \cellcolor{blue!8}\textbf{Land.} 
        & \cellcolor{blue!8}\textbf{Cele.} 
        & \cellcolor{blue!8}\textbf{Logo} 
        & \cellcolor{blue!8}\textbf{Poly.} 
        & \cellcolor{blue!8}\textbf{Ord.} 
        & \cellcolor{blue!8}\textbf{Single} 
        & \cellcolor{blue!8}\textbf{Multi} 
        & \cellcolor{blue!8}\textbf{Avg.}    
        & \textbf{Avg.} \\
        
        \noalign{%
            \global\aboverulesep=\oldaboverulesep
            \global\belowrulesep=\oldbelowrulesep
        }
        \midrule
		Grounding-Dino\cite{oneREC_liu2024grounding} & --- & 61.3 & 52.7 & 31.5 & 61.4 & 8.2 & 58.3 & 25.2 & 42.7 & 85.1 \\
		Qwen2-VL\cite{mllm_qwen2vl} & 7B & 68.4 & 59.3 & 21.7 & 46.8 & 9.2 & 16.7 & 16.8 & 34.1 & 88.4 \\
		Qwen2.5-VL\cite{mllm_qwen2.5vl} & 7B & 85.0 & 85.6 & 27.6 & 64.5 & 15.9 & 61.8 & 28.1 & 52.6 & 91.2 \\
		Qwen2.5-VL\cite{mllm_qwen2.5vl} & 32B & 85.5 & \textbf{91.4} & 31.5 & 61.4 & \textbf{46.7} & 61.8 & \textbf{37.4} & 59.4 & 93.5 \\
		LLaVA-OV-1.5\cite{mllm_llava_onevision1.5} & 8B & 70.5 & 75.7 & 7.6 & 59.1 & 6.7 & 19.1 & 8.7 & 35.3 & 86.9 \\
		GLM-4.6V\cite{mllm_glm4.5/4.6v_2025} & 9B & 86.5 & 83.5 & 55.9 & 68.2 & 24.1 & 72.9 & 21.1 & 58.9 & 90.1 \\
		Keye-VL-1.5\cite{mllm_keye} & 8B & 67.9 & 77.0 & 12.4 & 64.4 & 23.6 & 26.4 & 13.3 & 40.7 & 84.5 \\
		MiniCPM-V-4.5\cite{mllm_minicpmv45} & 8B & 78.8 & 84.5 & 33.5 & 54.9 & 24.6 & 53.5 & 11.8 & 48.8 & 87.2 \\
		InternVL3.5\cite{mllm_internvl3.5} & 8B & 72.5 & 77.0 & 2.9 & 52.1 & 35.4 & 36.0 & 9.4 & 40.8 & 85.6 \\
		Mistral-3\cite{mllm_mistral_small_3_2} & 8B & 8.2 & 24.8 & 0.7 & 4.6 & 2.3 & 0.7 & 0.2 & 5.9 & 32.1 \\
		MiMo-VL-RL\cite{mllm_mimovl2025} & 7B & 60.7 & 70.5 & 8.9 & 65.5 & 15.1 & 27.7 & 15.0 & 37.6 & 83.0 \\
		Qwen3-VL\cite{mllm_qwen3-vl} & 2B & 83.1 & 89.1 & 34.9 & 63.5 & 12.3 & 66.6 & 29.6 & 54.2 & 90.8 \\
		Qwen3-VL\cite{mllm_qwen3-vl} & 4B & 87.1 & 89.0 & 56.8 & 68.6 & 15.9 & 75.2 & 33.7 & 60.9 & 92.4 \\
		Qwen3-VL\cite{mllm_qwen3-vl} & 8B & \textbf{88.6} & 90.8 & \textbf{57.0} & \textbf{71.4} & 27.7 & \textbf{74.7} & 35.5 & \textbf{63.7} & \textbf{94.1} \\
        \bottomrule
    \end{tabularx}
\end{table*}

\section{Experiments}
To comprehensively evaluate proposed OpenRef benchmark and MCC approach, we conduct experiments across 14 representative models, including MLLMs and open-vocabulary models, such as Qwen2-VL\cite{mllm_qwen2vl}, Qwen2.5-VL\cite{mllm_qwen2.5vl}, Qwen3-VL\cite{mllm_qwen3-vl}, LLaVA-OneVision-1.5\cite{mllm_llava_onevision1.5}, GLM-4.6V\cite{mllm_glm4.5/4.6v_2025}, Keye-VL-1.5\cite{mllm_keye}, MiniCPM-V\cite{mllm_minicpmv45}, InternVL-3.5\cite{mllm_internvl3.5}, Mistral-3\cite{mllm_mistral_small_3_2}, MiMo-VL-RL\cite{mllm_mimovl2025} and Grounding Dino\cite{oneREC_liu2024grounding}. Experimental results demonstrate that OpenRef poses substantially greater challenges than RefCOCO, with all models exhibiting notable performance degradation, underscoring its effectiveness in assessing real-world grounding capabilities. Further, MCC is proved to be an intuitive but effective reasoning strategy to substantially improve REC models in OpenRef without retraining.
\subsection{Experimental Results}
\textbf{Main results on grounding accuracy based on F1 score.} We evaluate grounding accuracy of REC models using the proposed OpenRef benchmark. The quantitative results measured by F1 score are detailed in  \cref{tab:main_results_comparison}. Qwen3-VL series outperform other REC models in OpenRef, with the 8B variant outperforming all competitors across most subsets, and even 4B model surpassing much larger dense models like Qwen2.5-VL-32B. We observe that while specialist models like Grounding DINO excel in generic grounding, MLLMs exhibit a decisive advantage in knowledge-intensive categories such as Celebrities and Landmarks, leveraging their extensive pre-trained world knowledge. However, a common bottleneck remains in \textit{Ordinal}, \textit{Multi-target} and \textit{Logo} reasoning, where nearly all models experience significant performance degradation. Furthermore, despite their strong visual recognition capabilities, generalist models like Mistral-3 struggle to produce precise coordinates.
\begin{table*}
	\centering
	\fontsize{7pt}{7.5pt}\selectfont 
	\renewcommand{\arraystretch}{1.2}
	\setlength{\tabcolsep}{0.5pt} 
	\caption{Negative relative rejection evaluation on OpenRef. Best results are in \textbf{bold}.}
	\label{tab:N3R results}
	
	\begin{tabularx}{\textwidth}{@{} l c *{6}{C} @{}}
		\toprule
		\textbf{Model} & \textbf{Size} & \textbf{Land.} & \textbf{Cele.} & \textbf{Logo} & \textbf{Ordinal} & \textbf{\mbox{Targ. Mul.}} & \textbf{Avg.} \\
		\midrule
		Grounding-Dino\cite{oneREC_liu2024grounding} & --- & 40.0 & 36.8 & 38.3 & 70.6 & 50.0 & 47.1 \\
		Qwen2-VL\cite{mllm_qwen2vl} & 7B & 96.4 & \textbf{100} & 87.7 & \textbf{84.6} & \textbf{85.6} & \textbf{90.9} \\
		Qwen2.5-VL\cite{mllm_qwen2.5vl} & 7B & 29.0 & 13.8 & 4.4 & 4.2 & 19.0 & 14.1 \\
		Qwen2.5-VL\cite{mllm_qwen2.5vl} & 32B & 79.8 & 45.7 & 44.8 & 6.9 & 12.5 & 37.9 \\
		LLaVA-OV-1.5\cite{mllm_llava_onevision1.5} & 8B & 100 & 94.5 & 74.7 & 27.8 & 78.6 & 75.1 \\
		GLM-4.6V\cite{mllm_glm4.5/4.6v_2025} & 9B & 100 & 90.3 & \textbf{94.0} & 41.4 & 67.8 & 78.7 \\
		Keye-VL-1.5\cite{mllm_keye} & 8B & 96.4 & 74.0 & 83.6 & 28.3 & 69.3 & 70.3 \\
		MiniCPM-V-4.5\cite{mllm_minicpmv45} & 8B & 98.5 & 94.0 & 83.8 & 22.8 & 68.8 & 73.6 \\
		InternVL3.5\cite{mllm_internvl3.5} & 8B & 68.9 & 52.3 & 42.8 & 56.6 & 34.6 & 51.0 \\
		Mistral-3\cite{mllm_mistral_small_3_2} & 8B & 98.5 & 82.3 & 87.3 & 7.5 & 79.0 & 70.9 \\
		MiMo-VL-RL\cite{mllm_mimovl2025} & 7B & 22.1 & 20.5 & 15.6 & 27.7 & 17.8 & 20.7 \\
		Qwen3-VL\cite{mllm_qwen3-vl} & 2B & 92.7 & 60.9 & 23.1 & 4.5 & 9.2 & 38.1 \\
		Qwen3-VL\cite{mllm_qwen3-vl} & 4B & 92.2 & 60.8 & 23.4 & 4.3 & 8.6 & 37.9 \\
		Qwen3-VL\cite{mllm_qwen3-vl} & 8B & \textbf{100} & 92.1 & 90.1 & 58.0 & 82.6 & 84.6 \\
		\bottomrule
	\end{tabularx}
\end{table*}

\textbf{Main results based on negative relative rejection reliablity (N3R).} We evaluate relative rejection reliablity on negative expressions by using the proposed N3R metric. According to the robustness evaluation results in \cref{tab:N3R results}, we observe that specialized REC model i.e., Grounding DINO demonstrates limited rejection capabilities in proper nouns such as \textit{Landmark}, \textit{Celebrity} and \textit{Logo}, with significantly lower rejection scores compared to most MLLMs. MiMo-VL-RL and Qwen2.5-VL-7B also perform poorly in hard negative expressions. Generalist MLLM Mistral-3 exhibit surprisingly competitive performance on par with other MLLMs. Furthermore, within the Qwen2.5-VL and Qwen3-VL series, there is a clear scaling effect where the rejection reliability of smaller models is notably inferior to the larger variant.

\newcommand{\gain}[2]{\textbf{#1} \textcolor{red}{\textbf{\scriptsize($\uparrow$#2)}}}
\newcommand{\loss}[2]{\textbf{#1} \textcolor{gray}{\textbf{\scriptsize($\downarrow$#2)}}}
\begin{table*}
	\centering
	\fontsize{7pt}{7.5pt}\selectfont 
	\setlength{\tabcolsep}{1.2pt} 
	\renewcommand{\arraystretch}{1.4} 
	\caption{Effect of MCC to improve the grounding accuracy (F1) of existing REC models w.r.t. positive expressions, as a plug-and-play tool. Gains are marked in red.}
	\label{tab:ablation_mcc_onedecimal}
	
	\begin{tabularx}{\textwidth}{@{} l *{7}{C} c @{}}
		\toprule
		\textbf{Model} & \textbf{Land.} & \textbf{Celebri.} & \textbf{Logo} & \textbf{Poly.} & \textbf{Ordinal} & \textbf{Single} & \textbf{Multi} & \textbf{Avg.} \\
		\midrule
		Qwen3-VL-2B      & 83.1 & 89.1 & 34.9 & 63.5 & 12.3 & 66.6 & 29.6 & 54.2 \\
		+MCC  & 89.6 & 86.6 & 35.6 & 68.1 & 11.8 & 66.2 & 37.7 & \gain{56.5}{2.3} \\
		\midrule
		GLM-4.6V         & 86.5 & 83.5 & 55.9 & 68.2 & 24.1 & 72.9 & 21.1 & 58.9 \\
		+MCC     & 85.3 & 81.6 & 60.0 & 84.9 & 23.6 & 72.6 & 60.0 & \gain{66.9}{8.0} \\
		\midrule
		InternVL-3.5     & 72.5 & 77.0 & 2.9 & 52.1 & 35.4 & 36.0 & 9.4 & 40.8 \\
		+MCC & 78.2 & 77.0 & 14.8 & 62.1 & 37.5 & 49.5 & 17.0 & \gain{48.0}{7.2} \\
		\midrule
		LLaVA-OV-1.5     & 70.5 & 75.7 & 7.6 & 59.1 & 6.7 & 19.1 & 8.7 & 35.3 \\
		+MCC  & 69.4 & 69.9 & 8.1 & 63.2 & 5.1 & 20.2 & 8.5 & \loss{34.9}{0.4} \\
		\midrule
		Keye-VL-1.5      & 67.9 & 77.0 & 12.4 & 64.4 & 23.6 & 26.4 & 13.3 & 40.7 \\
		+MCC  & 73.6 & 79.7 & 9.8 & 68.0 & 29.7 & 27.0 & 16.0 & \gain{43.4}{2.7} \\
		\bottomrule
	\end{tabularx}
\end{table*}
\subsection{Evaluation of MCC on OpenRef}
To evaluate the effectiveness of the proposed Multi-task Consistency Checker (MCC), we conduct extensive experimental studies across a diverse range of MLLMs including Qwen3-VL-2B\cite{mllm_qwen3-vl}, GLM-4.6V\cite{mllm_glm4.5/4.6v_2025}, InternVL-3.5\cite{mllm_internvl3.5}, LLaVA-OneVision-1.5\cite{mllm_llava_onevision1.5} and Keye-VL-1.5\cite{mllm_keye}. As summarized in \cref{tab:ablation_mcc_onedecimal} and \cref{tab:ablation_fixed}, MCC yields substantial performance dividends on different MLLMs. MCC significantly enhances grounding accuracy particularly in multi-target scenarios. The most significant improvement is achieved in GLM-4.6V, with a F1 performance gain of 8.0\%. Furthermore, MCC also yields significant improvements in scenarios involving the rejection of negative expressions, improving model reliability by rectifying hallucinations and filtering false positives. The performance improvement is particularly pronounced for small models, with MCC substantially enhancing the rejection capability of Qwen3-VL-2B over 58.0\% and 53.8\% for absolute metric $N_{abs}$ and relative metric N3R, respectively. These results underscore MCC as a robust, training-free and plug-and-play inference strategy capable of harmonizing between-task inconsistencies, bolstering grounding accuracy and reducing REC hallucinations of MLLMs in unconstrained open-world environments.
\begin{table*}[t]
	\fontsize{7pt}{7.5pt}\selectfont 
	\caption{Effect of MCC to improve relative rejection reliability (absolute metric vs. N3R) of existing REC models w.r.t. negative expressions. Gains are marked in red.}
	\label{tab:ablation_fixed}
	\centering
	\setlength{\tabcolsep}{0pt} 
	\renewcommand{\arraystretch}{1.4} 
	\begin{tabularx}{\textwidth}{@{} l *{10}{C} C @{} C @{}} 
		\toprule
		\multirow{2}{*}{\textbf{Model}} & \multicolumn{2}{c}{\textbf{Land.}} & \multicolumn{2}{c}{\textbf{Cele.}} & \multicolumn{2}{c}{\textbf{Logo}} & \multicolumn{2}{c}{\textbf{Ordinal}} & \multicolumn{2}{c}{\textbf{Com. Scen.}} & \multicolumn{2}{c}{\textbf{Avg.}} \\
		\cmidrule(lr){2-3} \cmidrule(lr){4-5} \cmidrule(lr){6-7} \cmidrule(lr){8-9} \cmidrule(lr){10-11} \cmidrule(lr){12-13}
		& $N_{abs}$ & $N3R$ & $N_{abs}$ & $N3R$ & $N_{abs}$ & $N3R$ & $N_{abs}$ & $N3R$ & $N_{abs}$ & $N3R$ & $N_{abs}$ & $N3R$ \\
		\midrule
		Qwen3-VL-2B & 91.7 & 92.7 & 56.1 & 60.9 & 12.1 & 23.1 & 0.0 & 4.5 & 4.0 & 9.2 & 32.8 & 38.1 \\
		+MCC        & 100.0 & 100.0 & 96.7 & 97.0 & 90.2 & 91.5 & 92.3 & 94.4 & 75.0 & 76.4 & \textbf{90.8} & \textbf{91.9} \\
		            & & & & & & & & & & & \textcolor{red}{\textbf{$\uparrow$58.0}} & \textcolor{red}{\textbf{$\uparrow$53.8}} \\
		\midrule
		GLM-4.6V    & 100.0 & 100.0 & 86.2 & 90.3 & 90.8 & 94.0 & 17.4 & 41.4 & 58.8 & 67.8 & 70.6 & 78.7 \\
		+MCC        & 100.0 & 100.0 & 83.3 & 87.8 & 88.0 & 90.4 & 4.1 & 23.0 & 58.4 & 63.7 & \textbf{66.8} & \textbf{73.0} \\
		            & & & & & & & & & & & \textcolor{gray}{\textbf{$\downarrow$3.8}} & \textcolor{gray}{\textbf{$\downarrow$5.7}} \\
		\midrule
		InternVL-3.5 & 66.3 & 68.9 & 36.4 & 52.3 & 38.1 & 42.8 & 48.7 & 56.6 & 29.2 & 34.6 & 43.8 & 51.0 \\
		+MCC        & 100.0 & 100.0 & 72.4 & 84.2 & 66.4 & 74.1 & 14.9 & 28.4 & 48.6 & 59.3 & \textbf{60.5} & \textbf{69.2} \\
		            & & & & & & & & & & & \textcolor{red}{\textbf{$\uparrow$16.7}} & \textcolor{red}{\textbf{$\uparrow$18.2}} \\
		\midrule
		LLaVA-OV-1.5 & 100.0 & 100.0 & 94.1 & 94.5 & 73.4 & 74.7 & 22.1 & 27.8 & 77.9 & 78.6 & 73.5 & 75.1 \\
		+MCC        & 99.5 & 99.7 & 98.7 & 99.4 & 84.5 & 91.9 & 95.4 & 97.7 & 90.6 & 95.3 & \textbf{93.7} & \textbf{96.8} \\
		            & & & & & & & & & & & \textcolor{red}{\textbf{$\uparrow$20.2}} & \textcolor{red}{\textbf{$\uparrow$21.7}} \\
		\midrule
		Keye-VL-1.5 & 93.3 & 96.4 & 59.0 & 74.0 & 78.0 & 83.6 & 4.1 & 28.3 & 55.6 & 69.3 & 58.0 & 70.3 \\
		+MCC        & 94.8 & 96.5 & 78.2 & 83.3 & 84.0 & 85.8 & 44.6 & 52.2 & 71.6 & 77.6 & \textbf{74.6} & \textbf{79.1} \\
		            & & & & & & & & & & & \textcolor{red}{\textbf{$\uparrow$16.6}} & \textcolor{red}{\textbf{$\uparrow$8.8}} \\
		\bottomrule
	\end{tabularx}
\end{table*}

\section{Conclusion}
In this paper, we proposed \textbf{OpenRef}, a comprehensive benchmark designed to bridge the gap between closed environments and complexities of open-world REC. By relaxing traditional single-target assumption and incorporating complex visual domains and referring expressions, OpenRef establishes a new frontier for assessing grounding accuracy and rejection reliablity. To ensure high-quality data construction, we developed a semi-automated annotation pipeline, leveraging MLLM to generate annotations while maintaining strict human oversight for semantic precision. We introduced F1 score to measure grounding accuracy w.r.t. positive expressions and N3R to assess REC hallucinations w.r.t. negative expressions in open-world settings. Furthermore, we proposed MCC, a training-free and plugged-played self-verification mechanism, to effectively improve grounding accuracy and eliminate hallucinations (false positives) at inference. 
Extensive experiments demonstrate that this work significantly advances the existing REC benchmarks and models, paving the way for open-world REC.

\section*{Acknowledgements}
This work was partially supported by National Natural Science Fund of China under Grants 92570110 and 62271090, Chongqing Natural Science Fund under Grant CSTB2024NSCQ-JQX0038, and National Youth Talent Project.

\bibliographystyle{splncs04}
\bibliography{reference.bib}

\end{document}